\RequirePackage{fix-cm}
\documentclass[twocolumn]{svjour3}          
\smartqed  
\usepackage{graphicx}
\usepackage{subcaption}
\usepackage{scrextend}
\usepackage[disable]{todonotes}

\begin{document}

\title{The AI Settlement Generation Challenge in Minecraft
}
\subtitle{First Year Report}

\titlerunning{GDMC - First Year Report}        

\author{Christoph Salge \and Michael Cerny Green \and Rodrigo Canaan \and Filip Skwarski \and Rafael Fritsch \and Adrian Brightmoore \and Shaofang Ye  \and Changxing Cao \and Julian Togelius  
}

\authorrunning{C. Salge et.al.} 

\institute{           Christoph Salge \at
              Adaptive Systems Research Group \\
              University of Hertfordshire\\
              \email{ChristophSalge@gmail.com}\\
           \and
C.S., M.C.G., R.C. and J.T. are at the \at
                Game Innovation Lab\\
                New York University 
}

\date{Received: date / Accepted: date}

\maketitle

\begin{abstract}
This article outlines what we learned from the first year of the AI Settlement Generation Competition in Minecraft, a competition about producing AI programs that can generate interesting settlements in Minecraft for an unseen map. This challenge seeks to focus research into adaptive and holistic procedural content generation. Generating Minecraft towns and villages given existing maps is a suitable task for this, as it requires the generated content to be adaptive, functional, evocative and aesthetic at the same time. Here, we present the results from the first iteration of the competition. We discuss the evaluation methodology, present the different technical approaches by the competitors, and outline the open problems. 
\end{abstract}

\section{Introduction}

The Generative Design in Minecraft (GDMC) competition is an artificial intelligence (AI) competition open to the general public. The GDMC competition is designed to encourage people to make AI programs with human-level performance in a computational creativity domain. While creativity is a central element of human intelligence, it is hard to tackle with most search or optimization based approaches, as it lacks a clear goal condition or utility function. The GDMC competition aims to fill this gap, and hopes to provide a framework that can help to address some of the fundamental challenges in procedural content generation, such as adaptivity to existing content. 

In this article we report on the progress of the competition after the first year, including the submitted entries and their evaluation. We will also briefly review the technical and organizational aspects of the competition, though for a detailed discussion of those we refer the reader to our earlier paper~\cite{Salge:2018:GDM:3235765.3235814}. Building on top of this previous paper, these are the three new topics covered in this paper: 

First, we discuss the evaluation methodology and present the numeric results of the first year. We take a closer look on how well our evaluation based on human judges works, and discuss our approach to grade entries in four different categories. Second, based on written accounts by the competitors of the first year's competition, we outline the different technical approaches to AI settlement generation in their submissions. We describe the algorithms used, and analyze the strength and weaknesses of the different entries, with comparisons to the state of the art in academic literature.  Finally, we outline some of the open problem that are yet to be addressed. We also introduce and discuss the main addition in the next iteration of the competition, the optional bonus challenge of ``Chronicle Generation.''

\section{The GDMC Competition}

The GDMC AI Settlement Generation Challenge is a competition where competitors write an AI program that can create an interesting settlement for an unseen Minecraft~\cite{game:Minecraft} map. It was introduced~\cite{Salge:2018:GDM:3235765.3235814} to encourage research into procedural content generation (PCG)~\cite{short2017procedural,katecompton2016,colton2012computational,shaker2016procedural,liapis2014computational} in games.

Minecraft is a popular open-world survival game where the player is dropped into a world consisting of cubes. The play largely consists of mining existing blocks and building tools and structures with them. While the game has the actual goal of defeating the Ender Dragon, many players rather focus on building houses or settlements. This provides a rich baseline of examples for what humans can do in terms of settlement generation. Minecraft also contains a built-in settlement generator that constructs simple villages, giving us a technical baseline for algorithmic village generation. These two properties, together with the popularity of the game, led to us choosing it as the basis for our competition. 

We chose settlement generation as the first challenge in the Generative Design for Minecraft competition, not only because we want to see better village generation in Minecraft, but because it also provides an interesting scientific challenge in terms of adaptive and holistic PCG. Unlike clean slate generation, the task is adaptive: the generated settlement has to fit or adapt to a range of existing maps. This relates closely to the somewhat understudied concept of appositional reasoning in AI design, i.e. creating a design appropriate to a given scenario with ill defined goals \cite[Chapt.~8.]{smith2012mechanizing}. This contrasts our challenge with most of the existing challenges in the Game AI domain, and moves it closer to the domain of computational creativity \cite{colton2012computational}, while still keeping the overall format of a competition. The GDMC challenge is also concerned with what is known as \emph{holistic} PCG, because it requires the algorithm to build one artifact that fulfills a range of very different requirements at the same time. Most existing  PCG~\cite{shaker2016procedural}, in contrast, aims for a divide-and-conquer approach, where different elements or aspects are generated independent from each other. Orchestrating different kinds of content~\cite{liapis2018orchestrating,antoniosliapis2015} is an active research topic with several open questions that successful entries to this challenge could potentially answer, or at least provide a test bed for.

We translated the different aspects of settlement generation: adaptivity to the existing map, providing functionality, evocative narrative, and basic aesthetics into four scoring categories, and challenged our contestants to create an AI program that can build a Minecraft settlement that addresses all of them at the same time. The scoring categories are:
\begin{description}
\item[Adaptation:] Settlements are shaped by their environment and shape it in return.
\item[Functionality:] Settlements provide functionality and affordances to people, or in the case of games, to players and NPCs.
\item[Evocative Narrative:] Settlements tell stories about the people who made them and the history that shaped them. 
\item[Aesthetics:] Settlements are intuitively designed in accordance with basic design principles, such as scale, proportion, etc. 
\end{description}
These four scoring categories form the basis for human evaluation, which we will discuss in more detail in the next section. More details about the challenge in general can be found on our website\footnote{http://gendesignmc.engineering.nyu.edu/} and in our previous paper~\cite{Salge:2018:GDM:3235765.3235814}. We also provided a framework based on MCEdit~\cite{mcedit} and an example AI program to get our participants started. The framework itself provides functionality for reading the Minecraft map format and allows the competitors to treat the map as a 3D array. Participants submit code that can interact with the map via methods that read or write a blocks at a specific 3d position. 

\section{Related Work}

The earliest games that used procedural content generation date to the early eighties. Games such as \emph{Rogue} and \emph{Elite} used algorithmic means to generate the gameworld during runtime, in order to save memory space and developer effort. These early explorations were followed by extended use of PCG in a large variety of games for a number of reasons, including making new types of gameplay possible and the aesthetics of particular content generation algorithms. Some games, including large franchises such as \emph{Diablo} and indie favorites such as \emph{Spelunky}, are entirely dependent on the procedural generation of game levels during runtime.

Research on procedural content generation has been aided by the existence of a number of challenges, such as competitions, which allow for a way of more or less directly comparing content generation methods and solutions. The first PCG competition in an academic setting was most likely the level generation track of the Mario AI competition; here, competitors procedurally generated levels for a version of the classic platformer Super Mario Bros~\cite{shaker20112010}. These were evaluated by having human players play pairs of levels and indicate which one they liked best.

A similar setup is used for the level generation track of the General Video Game AI competition~\cite{khalifa2016general}. Unlike for the Mario AI level generation track, the generators here are not supposed to generate levels for a particular game; instead, they have to generate levels for any game that is given to them (as specified in a particular game description language). Another game-based AI competition with a level generation track is the Angry Birds competition, where the level generation track challenges competitors to submit generators that can create interesting Angry Birds levels with an appropriate level of difficulty~\cite{stephenson20182017}. 
 
Compared to these competitions, the Minecraft Settlement Generation Competition is more open-ended. Like the other PCG competitions, it is judged by humans. However, Minecraft is considerably more open-ended than Super Mario Bros, Angry Birds or the GVGAI games, which are all linear and have straightforward win and loss conditions. Minecraft settlement generators do not start with a blank slate, but have to adapt to the maps they are given. This makes it the first competition for adaptive and holistic PCG.

\section{Evaluation and Results}
\subsection{Map Selection}

\begin{figure*}
    \begin{subfigure}[b]{0.33\textwidth}
        \includegraphics[width=\textwidth]{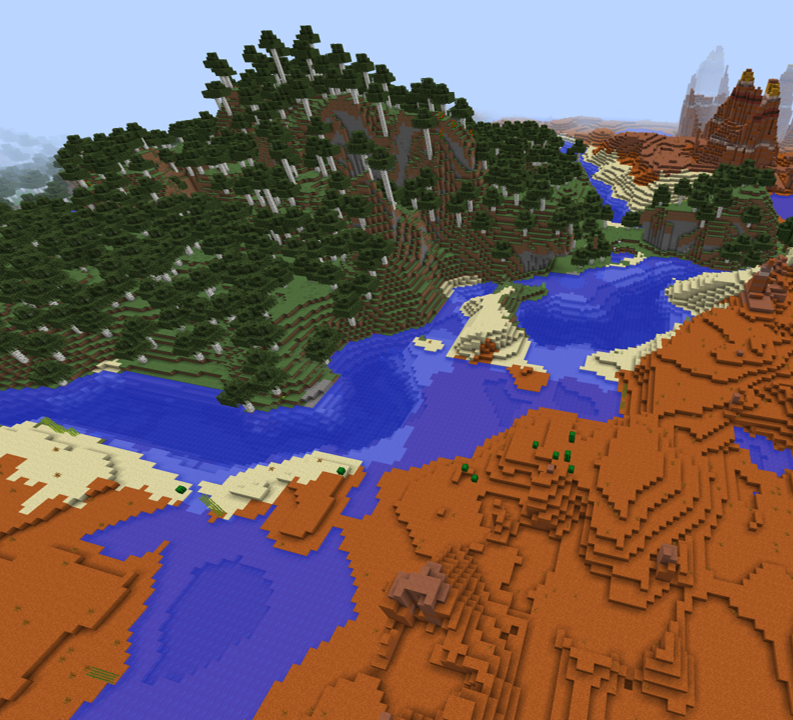}
        \caption{Map 1}
        \label{fig:map1}
    \end{subfigure}
    \begin{subfigure}[b]{0.33\textwidth}
        \includegraphics[width=\textwidth]{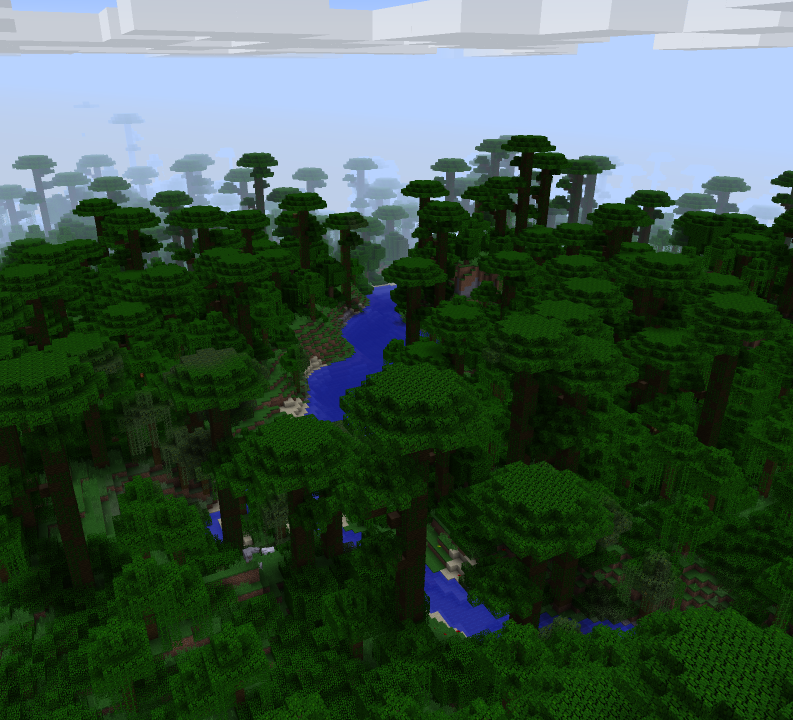}
        \caption{Map 2}
        \label{fig:map2}
    \end{subfigure}
    \begin{subfigure}[b]{0.33\textwidth}
        \includegraphics[width=\textwidth]{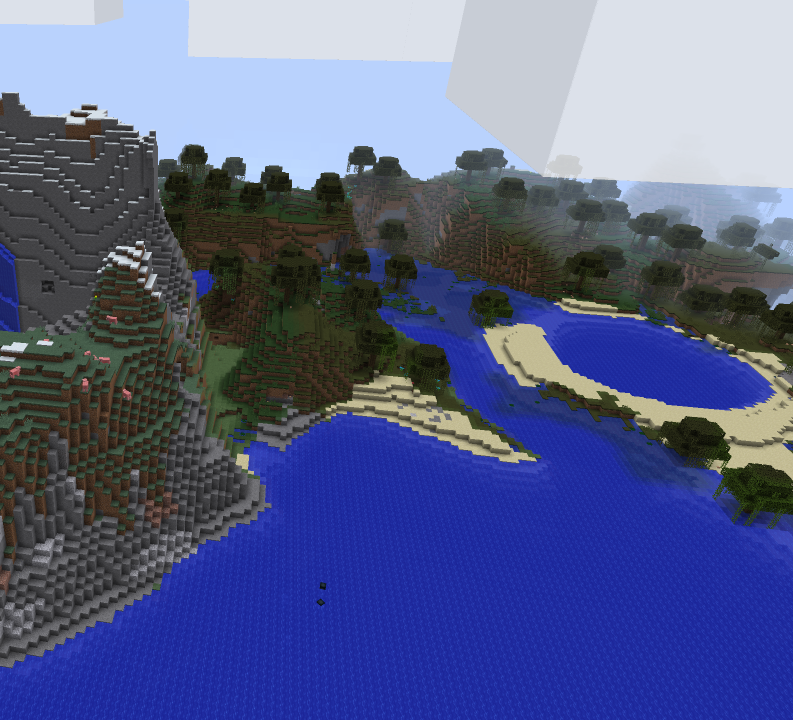}
        \caption{Map 3}
        \label{fig:map3}
    \end{subfigure}
\caption{The three maps used for the evaluation of the algorithms. }
\label{fig:maps}       
\end{figure*}

The first round of the competition ran until the end of June, 2018, with a total of 4 submissions. This section outlines how these entries were evaluated, present the actual results, and discuss some issues surrounding our evaluation methodology.

Three maps, unseen by any competitor, were created specifically for evaluating the generators. Part of the challenge was to see how the algorithms dealt with these unseen maps. Each map was generated with the standard world generator that comes packaged with Minecraft (Version 1.12). Aerial views of the three maps can be seen in Fig.~\ref{fig:maps} and downloads of the maps are available on the website.
\footnote{http://gendesignmc.engineering.nyu.edu/files/testMaps.zip}

These maps were chosen based on several principles. We wanted to have maps with increasing difficulty. Everyone should be able to build something reasonable for the first map while the third map should be challenging for everyone. Collectively, the maps should provide a range of different features so that the generators can showcase adaptability to a range of biomes, terrains, etc. Finally, we decided to only use maps generated by the game's built-in world generator, so the competitors would have a rough idea of what to expect. This last restriction may be relaxed in the future to include maps with hand-designed features, or maps generated with alternative world generators, such as those provided by mods.

The official competition maps can be seen in Fig.~\ref{fig:maps}. Map 1 is a relatively level map, with a river bisecting the terrain. The interesting feature here is the presence of two biomes, mesa on one side of the river and woodland on the other. Map 2 is a jungle biome map with another river running through the middle, and a small coastal section. The additional challenge of this map is a very dense tree cover that forces the algorithm to remove trees to create space for buildings. The third map is an island with the extreme hills biome, which provides extreme changes in elevation and is generally very challenging to build on due to the lack of flat terrain. It is also a terrain that is very hard for player to traverse, which raises a challenge of providing structures, such as stairs, train tracks, etc, that provide mobility and access to the player. 

\subsection{Evaluation by Judges}

The next step was to apply the 4 submitted algorithms to each of the 3 competition maps, generating 12 maps with settlement structures. These maps, along with the algorithms that created them, are also available on our website\footnote{http://gendesignmc.engineering.nyu.edu/}. Those maps with settlements were then send out to our team of expert judges for evaluation. The judges were recruited mainly from our advisory board and contained experts with various different backgrounds, including game designers, AI and PCG-focused scientists, urbanists, architects and Minecraft modders. See section~\ref{AB} in the appendix for details. We asked the judges to walk through each of the maps from a player perspective \footnote{The maps were actually in creative mode, so our judges would not be blown up by enemies} and then score each entry based on four criteria: adaptability, functionality, evocative narrative and aesthetics. For each of the criteria, the judges were given a non-exhaustive list of questions that should illustrate the criteria. Note that we repeatedly emphasized, to both the judges and the competitors, that the evaluation criteria are there as illustrating examples, and that we would rely on the judges interpretation and application of the overarching criteria to evaluate the submission. Subsequently, the kept the instructions relatively brief and generic, relying on the judges expertise to translate them as appropriate. The list can be found in our previous paper \cite{Salge:2018:GDM:3235765.3235814} or the supporting material. The judges were asked to score each category between 0 and 10, with the following instructions:
\begin{labeling}{6 -- 9}
    \item [0] the resulting design shows no consideration of that particular criterion at all. 
    \item [1 -- 4] there are some aspects in which the criterion is addressed.
    \item [5] this is comparable to a naive human. At this point, one would not be surprised if this was built by a human. 
    \item [6 -- 9] an expert-level human performance, over a longer time, possibly a group effort. So, we are talking about a group of city planners and architects designing a Minecraft settlement over the course of a year. The higher end of the point scale here should mean a work that would possibly win a design prize.
    \item[10] superhuman performance - this is so good, it would be surprising if this could be even generated by a dedicated group of expert humans. 
\end{labeling}

We provided this detailed list of what the different scores mean to somewhat anchor the scores. This should give us quantitative way to talk about a general improvement in submission, if the overall or average score of participants goes up in subsequent years. 

\subsection{Results}

Table~\ref{tab:1} shows the averaged scores from eight judges. Entry 1 won the competition with an average score of 4.38. The judges also gave detailed comments on the entries, which were sent to the participants. These informed our descriptions of the strength and weakness of the different entries in the next section. We can see from the results that entry 1 outscores the competition mainly in the adaptability and functionality category. This is not surprising, as communication with the contestants about their entries indicated that the main challenge they tried to address was the placement of buildings and roads. The distribution of the scores is narrower for the narrative and aesthetics section. The overall opinion of the judges in this area was that the aesthetics were satisfactory overall, if lacking in variability, while the category of evocative narrative was not really addressed by any entry. Overall, the scores are also mostly below 5, indicating that the resulting artifacts do not pass as human-made yet.

Since we wanted to have a challenge that looks at different aspects of PCG in general, and settlement generation in particular, we selected a multi-dimensional, scalar evaluation, rather than a preference comparison~\cite{togelius2016run}. We also chose a higher granularity than advised by literature, to account for both sub-human, human-like and trans-human performance. We computed the pairwise Pearson product-moment correlation coefficient \cite{pearson1895note}, which gives us an average value of 0.51, indicating a decent inter-rater reliability. It is harder to establish the validity of the measurement, but as the measurement is guided by a set of questions, it stands to reason that the measurement is related to the topics raised in those questions. We also note that 7 out of 8 judges agreed on the same overall winner.

\begin{table*}[t]
\centering
\caption{Results from the first GDMC competition}
\label{tab:1}  
\begin{tabular}{lllllll}
\hline\noalign{\smallskip}
Entry & Adaptability & Functionality & Narrative & Aesthetics & Overall Score & Rank \\
\noalign{\smallskip}\hline\noalign{\smallskip}
Entry1, Filip Skwarski &	5.42 &	4.71 &	3.17 &	4.21 &	4.38 & 1st \\
Entry2, Adrian Brightmoore &	2.33 &	2.21 &	2.13 &	3.54 &	2.55 & 3rd \\
Entry3, Changxing and Shaofang &	0.96 &	2.96 &	2.13 &	2.75 &	2.20 & 4th \\
Entry4, Rafael Fritsch &	2.81 &	3.25 &	2.46 &	3.79 &	3.08 & 2nd \\
\noalign{\smallskip}\hline
\end{tabular}
\end{table*}

\subsection{Lack of Quantitative Analysis}

One challenge with this particular competition is the difficulty of providing a quantitative analysis. For one, we specifically did not want to define any quantitative criteria for the competition, as we are concerned that this would lead to participants optimizing for said criteria, rather than trying to build a generator that addressed the problems of appositional design. Furthermore, we are also not sure that reasonable, quantitative criteria have been defined yet. While it would be possible to count block configuration in forms of bi- or trigrams, or calculate some form of spatial entropy \cite{togelius2010towards},
it is easily evident that neither a very high or very low value, i.e. perfect regularity or perfectly random blocks, is desirable as a good settlement design. 
Such metrics could be used to perform an expressive range analysis \cite{smith2010analyzing}, i.e. measuring for repeated creations how varied the generators are. While this is certainly an interesting question which could be explored in separate work, it does not necessarily provide an indication of the quality of the generators. Of the currently submitted generators, some are using a random seed, and as such are producing a different settlement when run repeatedly, while others are producing the same settlement every time, their only source of variation being the map as input. But as the main aim of the challenge is to find one design that fits well into the map this does not necessarily make the second kind of generator better or worse, as we are not interested in un-directed variation, but specifically in adaptation to existing variation. As a consequence, and also due to the limited capacity of our human judges, we only ran each map generator once for each map. 

In the future, it would be interesting to use this test set to find quantitative metrics that reflect the opinion of humans. One solution here might be to look for more embodied approaches that measure the actual interaction of an avatar with the generated artefact, i.e. the settlement. For now, we think that the most reasonable way to measure the quality and improvement of the generators is through human evaluation. As the main scientific aim of the competition is to elicit and discuss new approaches to the problems in adaptive and holistic PCG, we will also perform a qualitative analysis of the submitted generators. We did not perform a quantitative analysis of the existing approaches, as for example seen in \cite{Cooper2011}, but rather looked at how the competitors described their main ideas. 


\section{Submitted Generators}
This next section broadly explains the inner workings of the generators that were submitted to this round of the competition. Each subsection describes a specific entry, who made it, how it works, and what inspired the authors' design. Table~\ref{tab:2} displays a short summary of these results. We will then also provide a short analysis on how the different generators related to the state of the art, and examine if they provide approaches that could be generalized for scientific insight. 

\subsection{Entry 1, Filip Skwarski, 1st}

 \begin{figure}
     \centering
     \includegraphics[width=\columnwidth]{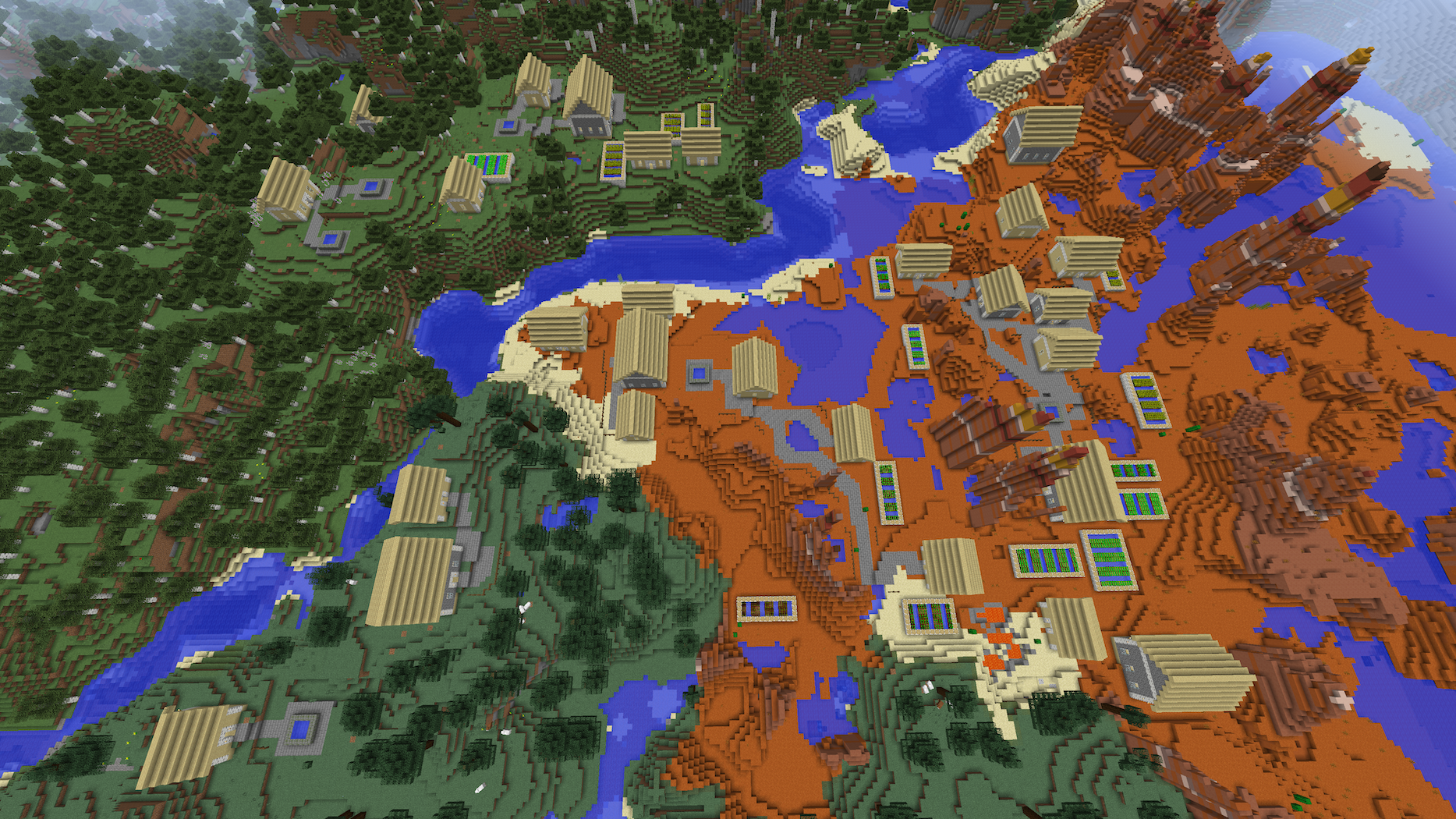}
     \caption{Aerial view of entry 1}
     \label{fig:entry_1}
 \end{figure}

This entry generates settlements in a bottom-up fashion by continuously filling the map with new structures. It is loosely inspired by the way human players build settlements: rather than following a fixed plan, they may construct and expand settlements in different ways depending on practical and aesthetic considerations, such as available resources and the shape of terrain. The algorithm models this process by ranking a large number of possible structures with regard to a number of features in order to select the optimal candidate to generate next, given previously generated structures and the shape of surrounding terrain.

Generation starts with a fixed list of structures that will be generated in order. In the submitted version the structures alternate between houses of two types, plazas and farms. Buildings are incrementally placed after evaluating 5000 random positions based on these three criteria:
\begin{enumerate}
    \item Elevation: how flat is the terrain around and in front of the building?
    \item Layout: does the building type fit in with the existing structures?
    \item Distance: how far from existing structures is the building?
\end{enumerate}
The Layout criteria refers to a list of soft constrains for each building, such as plazas wanting to be surrounded by houses and farms being aligned with existing structures. All structures are currently symmetrical, but their sizes are randomized within a certain parameter range. 

After the placement of each structure, a new road is added, connecting it with its closest neighbour. Roads are generated using the A* pathfinding algorithm~\cite{hart1968formal}.
In order to prevent them from ascending and descending hills diagonally, areas around 'corners' of hills are marked as impassable by default. Similarly, jumps of two or more blocks in height are considered impassable. This also means that in some cases, there is no possible connection and no road will be generated.

The product of this method is clusters of houses centered around fountains and connected by roads, as seen in Fig.~\ref{fig:entry_1}. These small communities sometimes have farms nearby. Often communities have roads built between them to assist in moving from one to another.
The algorithm considers discrepancies in elevation as well as impassable terrain (e.g. water, lava) when placing buildings, so areas of flat terrain are given priority when expanding the settlement. It also takes slopes into account when placing roads. Overall, the algorithm produces settlements that fit very well into the existing heightmap, which is particularly evident in Map 3, which features a lot of elevation.
It is also the only entry that properly clears trees, and not just removes those blocks of the tree that are in the way. It does not modify the landscape further, though. 
While the buildings are empty inside, they are lit from the outside and contain windows, so they may provide shelter when taken over by the player. The farms act as a source of food. The roads, which connect buildings, may help the player find their way back into the settlement.

The settlement is built incrementally and the choice of materials reflects the resources available (type of wood), which hopefully evokes the impression of ``growth'' or ``history'' behind it. The adaptation to terrain also means that the configuration of individual settlements differ from one another, e.g. a settlement built in the mountains will look distinct from a settlement built in an open plain. On the other hand, there is little variation in architectures in regards to the terrain or biome. 

From a scientific perspective, this winning generator utilizes several existing techniques common in academic literature and the industry. The created settlement is a composite of houses, farms and streets, which are all created with subroutines. The buildings vary a bit, and are expressed by a form of parameterized grammar, ensuring the vary with an aesthetically accepted design space (see \cite{shaker2016procedural}, chapters 3 and 5). The entry scores high in adaptivity, and employs several common strategies here. First, there is parameter-based adaptation, where the program detects specific discrete parameters, such as the dominant type of wood, and then makes changes based on that parameter, i.e. uses different wood blocks in buildings. The placement of the building is also a classical search-based approach where different candidate solutions for space are evaluated based on a fitness function~\cite{togelius2011search}. The two most interesting elements here are, first, the fact that later placements are influenced by earlier placement. This can be seen as a simulationist approach, as one might imagine that later placed houses were built later in the history of the village. The second interesting approach is the street placement that explicitly incorporated the embodied interaction of the player with the world. Street layouts are rejected based on steepness, which explicitly encodes what a player character can walk on or not. While the generator currently only encodes the standard walking model, it would be possible to extend this further, to look at adapting roads based on the interplay of terrain and different forms of mobility. 

\subsection{Entry 2, Adrian Brightmoore, 3rd}

 \begin{figure}
     \centering
     \includegraphics[width=\columnwidth]{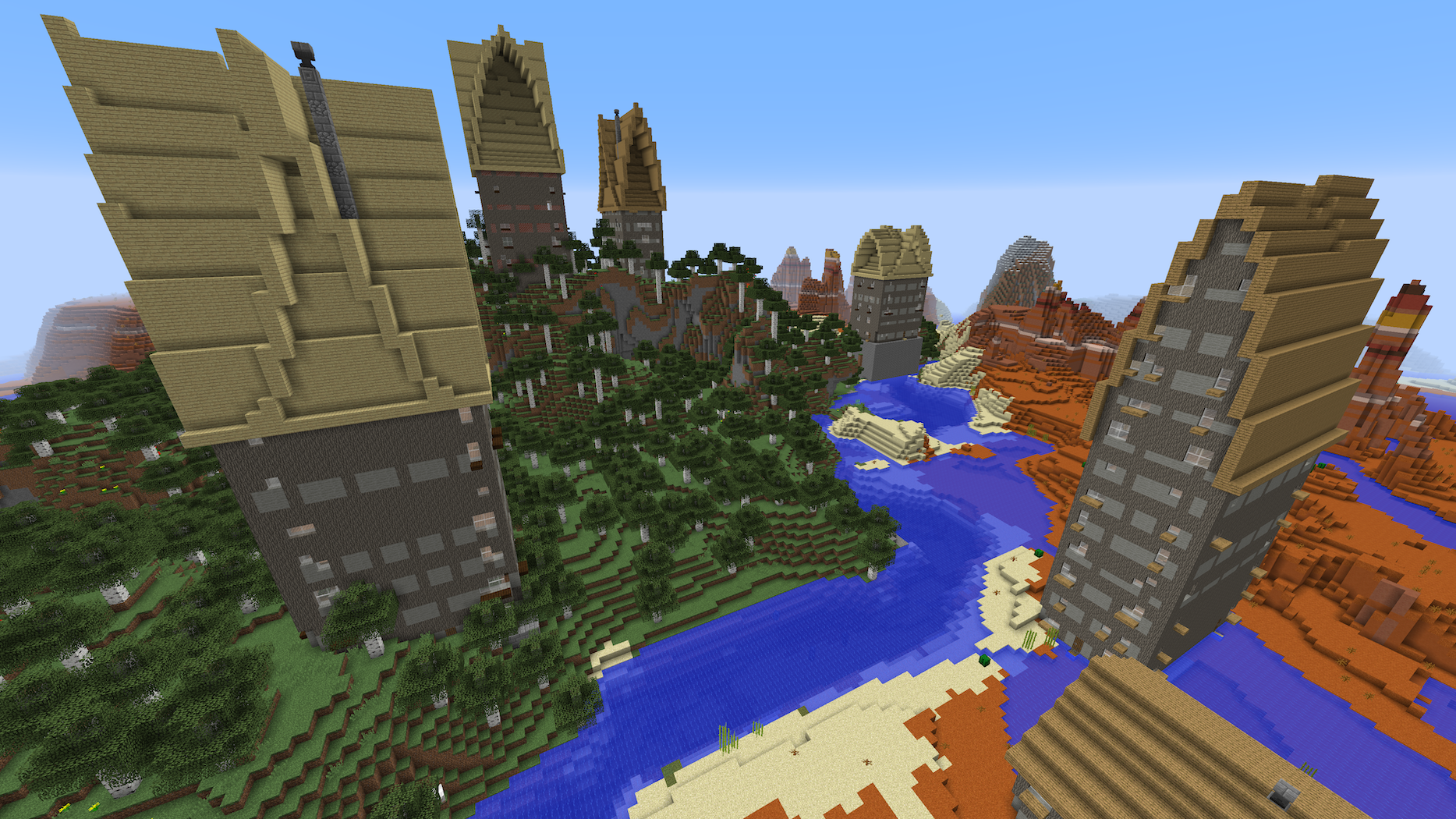}
     \caption{Aerial view of entry 2}
     \label{fig:entry_2}
 \end{figure}
 
\begin{figure}
     \centering
     \includegraphics[width=\columnwidth]{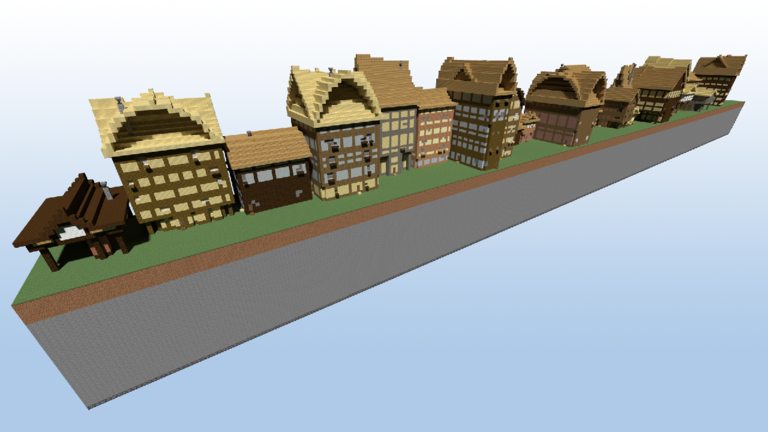}
     \caption{Examples of buildings generated to match the style of historic London.}
     \label{fig:london}
 \end{figure}

\begin{figure}
        \begin{subfigure}[b]{0.49\columnwidth}
        \includegraphics[width=\textwidth]{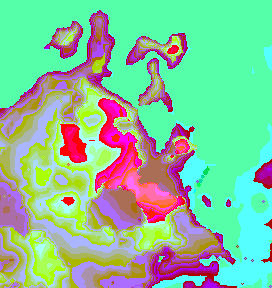}
        \caption{}
        \label{fig:2a}
    \end{subfigure}
    \begin{subfigure}[b]{0.49\columnwidth}
        \includegraphics[width=\textwidth]{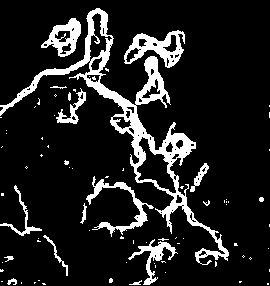}
        \caption{}
        \label{fig:2b}
    \end{subfigure}

\caption{Entry 2, Height map (left) and derived edge map (right) showing connected areas of gently sloped terrain}
\label{fig:heighmaps}       
\end{figure}

 This entry uses a two-step process and operates mostly on a 2d-representation of the map - more details can be found in a separate blog post~\cite{brightmoore}. It first analyses the height-map of the given map to determine the positions for all buildings. After reading in the height map, as seen in Fig.~\ref{fig:heighmaps}, it compares the height of neighbouring blocks. Any difference of two blocks or more, which is unjumpable in Minecraft, is considered an edge, and the area around it is then marked as too steep to build across, see Fig.~\ref{fig:2b}. Finding plots for the buildings is then simply about finding a plot of land that has no ``edge'' running through it. 
The buildings themselves are then generated by the building generator, see Fig.\ref{fig:london}. that was part of another project by the author, which attempted to create the greater London of the past \cite{london}.

The generator creates settlements which contain small collections of large buildings. Due to the buildings originally being intended for a city, the buildings seem bigger than expected for a small villages, but lack the supporting infrastructure to look like a city. The buildings tend to be widely spaced apart with little to no connecting roads or pathways to get from structure to structure.
The main adaptive feature of this entry was its height-sensitive placement of buildings, that attempts to avoid large cliffs, which would be dangerous to live next to.
Unlike other entries, this generator had internal building design. Structures contain multiple floors, stairs, windows, and are large enough to live in comfortably. Unfortunately, there is no pathway to get from structure to structure, however, and none of the structures are well lit.
Although there was not a dedicated narrative idea that shaped the generator, the author wanted to create a ``country living'' theme within the settlement, as though the people living here were on the frontiers of civilization. That effect was partially achieved by the scarcity of structures and the lack of infrastructure throughout.  

Again, from a scientific point the methods used here are well established. The unique aspect of this approach is mostly technical. The preprocessing of the obtained height map, using common computer graphic methods, could offer a good pathway to more efficiency. The winning entry 1 needed plenty of time to evaluate different house positions and possibly processed the same data repeatedly. Transforming the height data into a more effective representation and defining operators for suitability on this map might be a more efficient way forward. Furthermore, the integration of previously existing code demonstrates the capabilities of the chosen competition framework, as it allows for reuse of existing code, and as such can lead to more complex hierarchical solutions developed by several people, over several iterations. This could potentially allows us to also study the problem of how to orchestrate or exchange information between different parts of a PCG framework. 

\subsection{Entry 3, Shaofang Ye and Changxing Cao, 4th }


\begin{figure}
    \centering
    \includegraphics[width=\columnwidth]{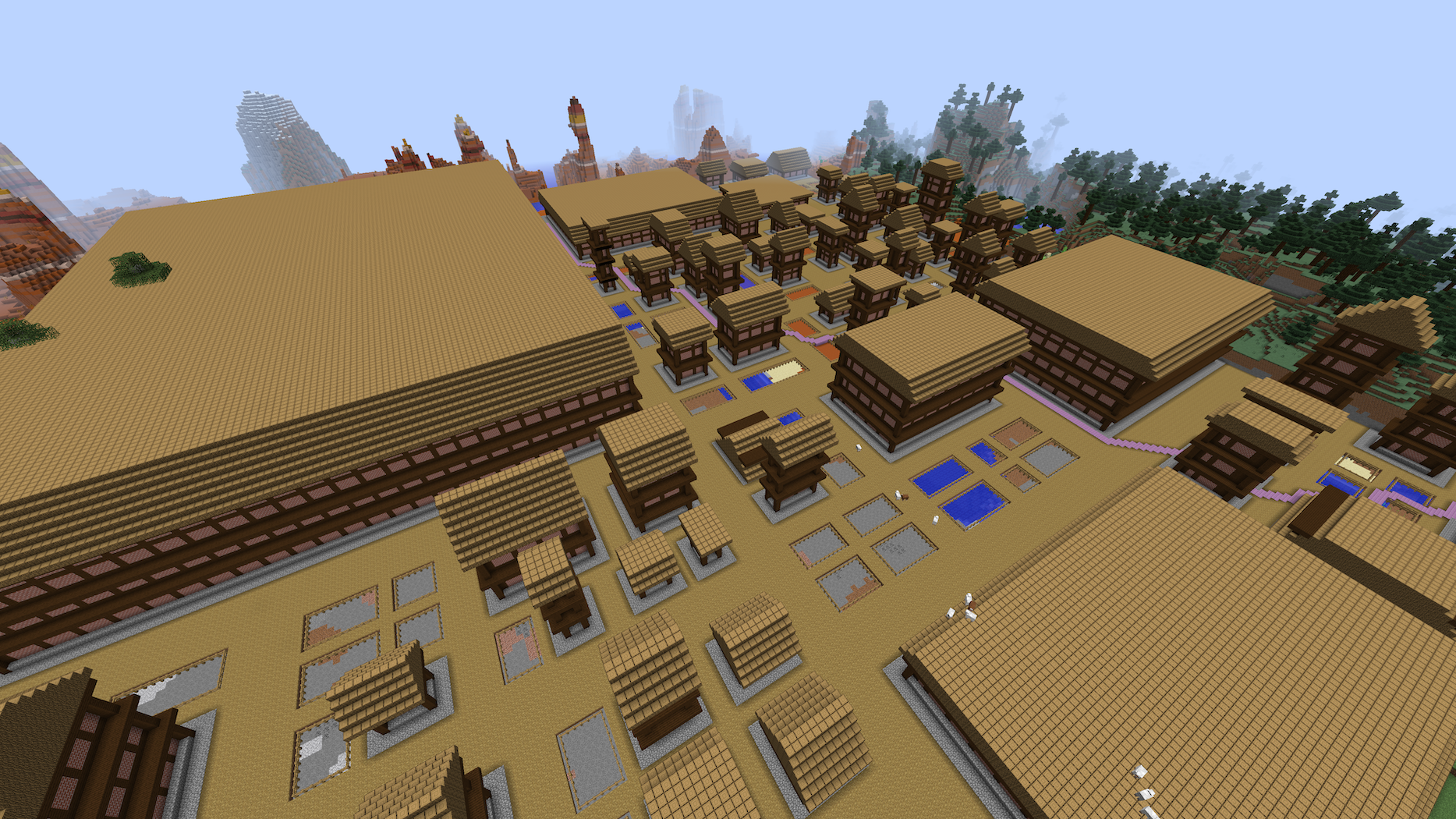}
    \caption{Aerial view of entry 3}
    \label{fig:entry_3}
\end{figure}

This program has a top-down approach that first levels the whole area and then partitions the space into plots for houses and fields. The approach was inspired by the grammar shaping method for procedural building generation called CGA shape~\cite{muller2006procedural}. The premise is to partition a three-dimensional space in a way that can be used to build modular cubic structures. The algorithm uses two core methods to achieve this: \textit{add} and \textit{cut}. The \textit{add} method takes a new cubic partition adjacent to the current cube and appends it to the space. The \textit{cut} method removes a formerly added partition. This method is used to decide ``yard'' placement, where a house and the adjacent road will be built. The yard is then further partitioned for both of these elements.

Structure generation is done by dividing the house into three areas: the roof, the walls, and the floor. All of these methods use \textit{cut} and \textit{add} as before, but with limited dimensionality (floors only on the x-z plane, walls on the y-z or y-x). Fences and farms are also part of the generative process and are constructed much in the same way. A single purple road traverses the city from one corner to the other. This was generated using the A* algorithm \cite{hart1968formal}, making sure to not overlap with buildings, farms, or other occupied space in the process.

The resulting settlements from this generator have a very structured feel. The entire area of the settlement is level, so walking around feels clean and easy. Even though no explicit road exists, other than a purple one which winds its way from one corner of the city to the other, the houses feel naturally positioned relative to one another, in a way that feels easy to get around.
The current implementation of this cubic partitioning process requires the land for the settlement to be completely flattened, therefore affording no adaptability. The materials in the houses are unrelated to the local materials available. The authors plan to allow more adaptability in future iterations. The function that builds the houses it well suited to build houses of different sizes, depending on the plot, which gives the houses a certain deal of variety. The range of size seems to be unbound though, which creates buildings that are strangely large, or builidings so small that there is only one square meter of internal space. 
In addition to housing, the settlements provide food in the shape of farms. The buildings themselves are smaller than what would be considered ``comfortable'' to live in, given the relative size of the player. There are also no lights to brighten the city at night, another addition the authors plan to add in future iterations.
The design of the structures and the addition of farmland give the sense that these settlements are inhabited by an agricultural people. The settlement feels like an outpost, or a small town that is slowly growing into a city.

From a scientific point this entry is a straightforward application of a top down, hierachical generator, that uses KD-trees~\cite{bentley1975multidimensional} for separation and a grammar based generator to build stuff in the nodes. It completely sidesteps the adaptivity issues by flattening the land, thereby basically making it a tabula rasa generator. 

\subsection{Entry 4, Rafael Fritsch, 2nd}

\begin{figure}
    \centering
    \includegraphics[width=\columnwidth]{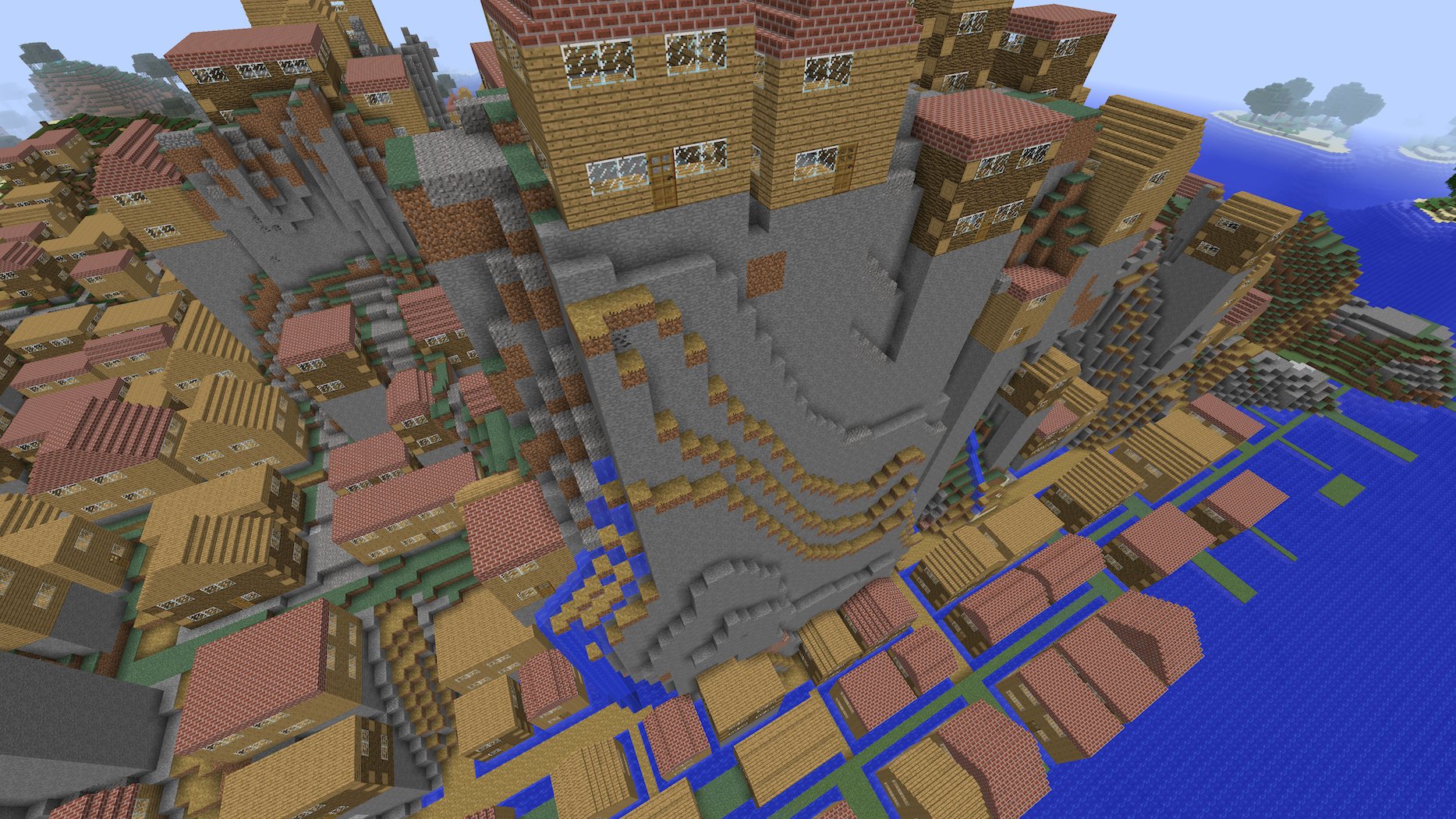}
    \caption{Aerial view of entry 4}
    \label{fig:entry_4}
\end{figure}

This entry was designed to adapt to the environment in terms material make-up. The first step of this process is to analyze the site. The occurrences of all block types at the site are counted. From this the frequencies of wood and stone types are determined, as they form the basis for choosing building materials in later steps. Next, the climatic conditions of the site are analyzed. By combining the frequency of biomes and information from the Minecraft Wiki the average temperature can be calculated. Further considering the amount of surface water and fertile ground (grass, dirt, etc) helps to decide what kind of vegetation and crops are appropriate for the building site in question. 

The entry also adapts to the ground level - and it initially caches the ground and surface levels for faster access. The ground level is defined by the highest solid block, ignoring things like vegetation and certain crafted objects. The surface is the same as the ground level, apart from water bodies, where the ground level is at the sea floor, whereas the surface level is at the sea level. Outside of water, ground and surface coincide. 

To organize the building sites, the generator then splits the map into rectangular plots. Currently this is done by repeated bisection of the available area, where each segment is split into two sub-segments, separated by a third section in the middle that is used to lay down roads. The size of the resulting plots is controlled by minimum and maximum constraints.
Before starting the structure-building process, the structure site in question (plus a small border) is cleared above the surface. Thus, newly-placed constructions are not obstructed by vegetation. The type of structure to be built on each plot is decided by considering its fitness for different 'builders'. For example, the fitness of the house builder in a plot depends on how central to other plots it is and whether it has a road next to it.
Roads are laid down in the previously designated plots. Their quality deteriorates towards the settlement border. In the centre stone is used, whereas further out you will find gravel and dirt tracks.
When building a farming plot, a random crop fitting the site's conditions is chosen. A season was associated with the site randomly or set by the user. The growth stage of the crops progresses from spring to autumn. In winter the fields lie fallow. Fields may be surrounded with wooden fences or hedges.

Houses are built in a modular approach. First the foundation is laid and the shell (floors and walls) is placed. After a roof is added, the interior design is addressed. This final step includes the placement of doors and windows, so they can be adapted to the room layout. However, room layout and furnishing is not implemented yet. The variety of building modules and adaptation of style and material to the site offers great potential for further improvement.

This method creates a city (Fig.~\ref{fig:entry_4}) similar in feel to a high density slum. Buildings are very close together, making one feel claustrophobic during exploration. Tiny, one-tile roadways separate houses in a way that feels closed-in and tightly-packed. This is only interrupted occasionally  by a small farm, but the number of farms can hardly be expected to support the apparent population of the settlement.
The generator selects wood and stone types for structure composition based on their abundance at the site. So if the land contains an abundance of mud and jungle trees, then the buildings will mainly be made of mud bricks and jungle wood. The crops used in farming plots are chosen based on biomes and soil/water availability. The presence of more soil and water, as well as being in a wetter biome like a jungle or grassland, results in more farms rather than less.
The generated structure provide simple housing for inhabitants. The farms provide food, but hardly enough to feed all of the city, so food must be imported from the outside.
The author of this entry admits that no narrative was designed into the generator explicitly. However, because of the aesthetics of the generator (as discussed above), the city feels economically poor. The houses are plain with little to no decoration and are very tightly packed together to fit as many inhabitants as possible.


\begin{table*}[t]
\centering
\caption{Summary of entries from the first GDMC Competition}
\label{tab:2}
\begin{tabular}{|l|l|l|l|}
\hline
Entry & Struture Placement            & Adaptability                        & Aesthetic                   \\ \hline
1     & incremental placement             & material, elevation, water          & clustered community village \\ \hline
2     & height-map and edge detection & elevation and water                 & a small country village     \\ \hline
3     & binary-space partitioning     & n/a                                 & wealthy urban               \\ \hline
4     & grid-based partitioning       & climatic, biome, seasonal, material & densely-populated urban     \\ \hline
\end{tabular}

\end{table*}

\section{Open Challenges}

In the previous section we describe how each entry addressed the competition evaluation criteria: Adaptability, Functionality, Narrative and Aesthetics. The main focus in this approach was on appropriate structure placement, and as Table~\ref{tab:2} outlines, there are basically two approaches, bottom up placement of single structures  (entries 1 and 2) , and a top down grid approach (entries 3 and 4). Adaptation beyond that was mostly in terms of categorical values, such as the type of wood present or the present biome. While the entries offer diverse approaches, utilizing a range of existing PCG methods, feedback from the judges also pointed out several aspects in which all entries are lacking. The overall lack of adaptive narrative led us to introduce the bonus challenge of chronicle generation, which we will discuss at the end of this section. Other open challenges, which if addressed can lead to improvement across multiple evaluation criteria, are also discussed below, pointing towards future opportunities of research.

\subsection{Building Variability}

All entries addressed building variability by generating houses in different sizes, shapes or materials. In some cases, the materials were chosen to adapt to the terrain, and in some cases other types of structures, such as farms and wells, were generated alongside houses to provide extra functionality.

However, the variability between the buildings themselves is not leveraged in a systematic way which could further improve the settlement - nor is the variety in the buildings adaptive to any underlying properties. Buildings that look similar could be placed together, giving rise to districts with individual aesthetic identities, would also hint at the history of the settlement's development. The exterior and interior of buildings could be used to suggest that building's functionality or to reflect its owner's personalities or social status. The placement of buildings could also relate to previous placement of a few key buildings to achieve some semblance of organic growth, such as a town growing outwards from a central church, castle or square. In addition, there is little-to-no variance in internal room generation. Since this competition ran, there has been research done in this area outside of a competition entry, which is meant to be used as an asset for future competitors~\cite{green2019organic}.

\subsection{Adaptation to Water}

\begin{figure}

       \includegraphics[width=\columnwidth]{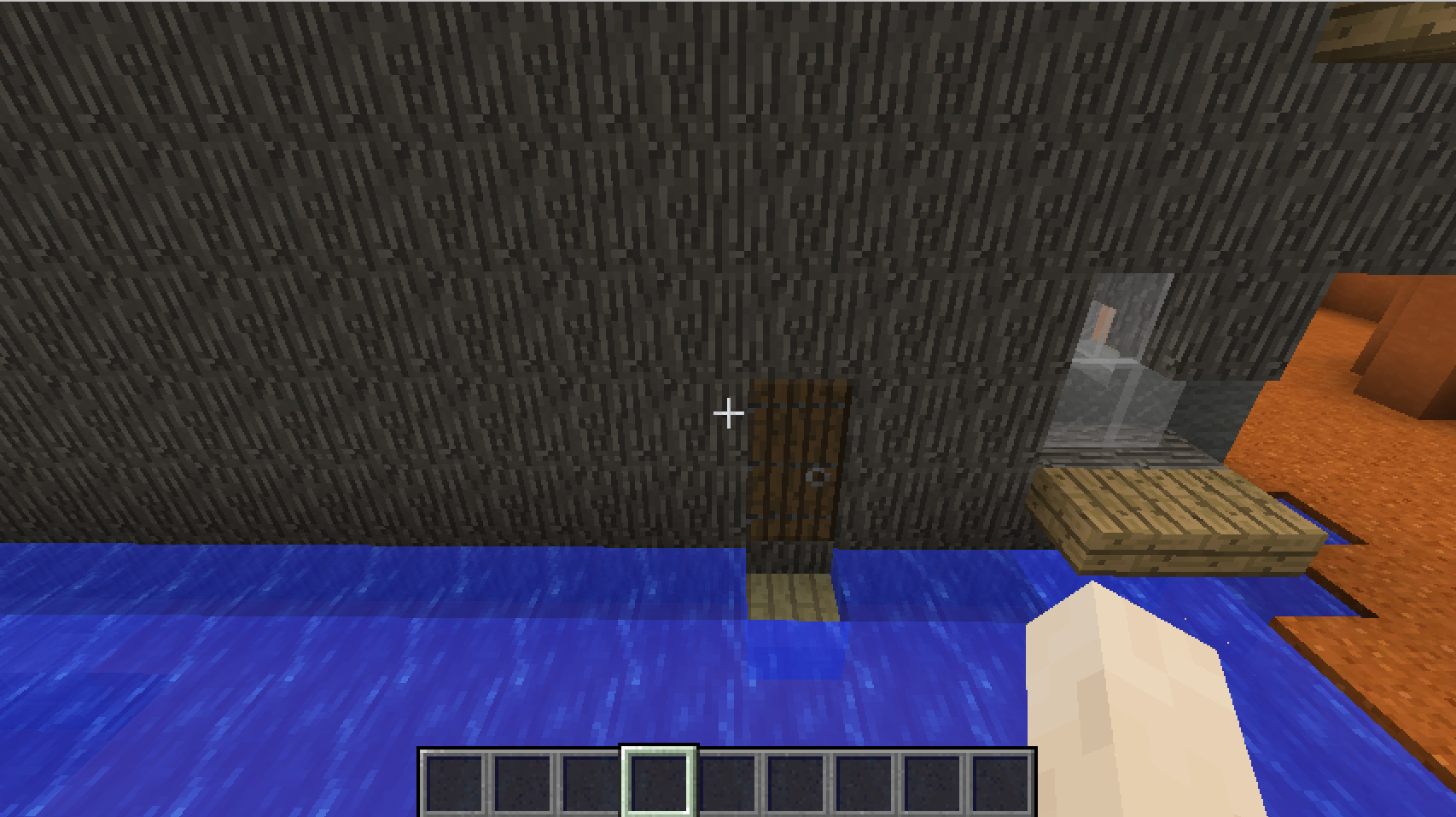}
       \caption{An inaccessible door leading into water}
       \label{fig:watera}
\end{figure}

\begin{figure}
        \includegraphics[width=\columnwidth]{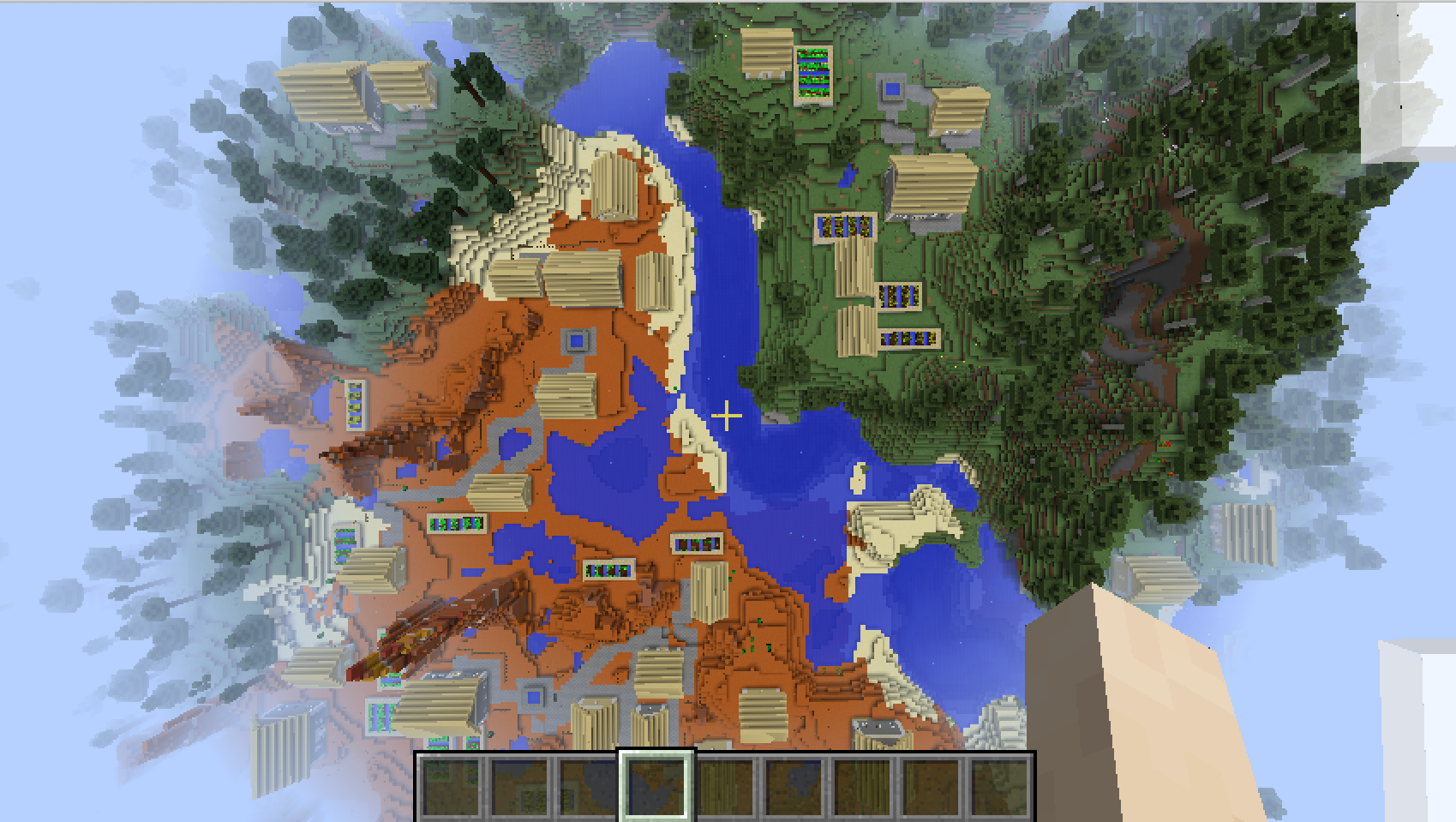}
        \caption{A city divided by a river, with no way to cross}
        \label{fig:waterb}

\end{figure}

A significant feature missing from all entries was adaptation to water. This was highlighted by several judges. Two of our maps had the settlement crossed by a river, while the third map had the settlement facing an ocean. Some entries had buildings placed unrealistically over water, or failed to take the proximity of water into account when placing doors, as seen in Fig.~\ref{fig:watera}. Even when building placement correctly took water into account, this resulted in a disconnected settlement, split by a river with no way to cross, as seen in Fig.~\ref{fig:waterb}.

The presence of bridges in such cases would greatly contribute to the Functionality and  Adaptability criteria. Rivers often also serve as natural borders between cities or countries in the real world, and could be used by entries to demarcate regions with a distinct feel. 

Bridges are not the only form of adaptation to water, however. If the body of water is large enough, such as an ocean or a big lake, it could feature ships, harbors, decks and so on. Houses on the waterside could also form a district of their own, with unique architecture and affordances. Currently settlements show little adaptation to water as they neither compensate for the existence of water, nor do they leverage the benefits of water.

\subsection{Big-Picture Adaptation}

While all submitted entries featured some adaptation to the provided map on a small scale (e.g. where to place structures, which materials and crops to use) none of the current approaches adapted on a larger scale, such as deciding to build a specific type of settlement or other global characteristics depending on the map's features. We call these big-picture adaptations.

For example, a settlement located in a narrow passage between mountains might have defensive purposes, and thus feature structures like castles, towers and walls. A dense jungle might be the ideal place to build an Elvish village, with strong ties to nature. A city by the sea, with lots of fishing vessels and a large harbor might have higher population density and more advanced architecture than a map with fewer resources.

Identifying and leveraging opportunities for these big-picture adaptations could lead to settlements that are tightly integrated with a specific map, potentially contribute towards scoring in all evaluation criteria and be important step for Holistic PCG in general.

\subsection{Data Driven Approaches}

Given the availability of data, a good way to approach the problem proposed by the competition might be to rely on existing examples of settlements generated by humans in Minecraft, or even real-world cities. As an example of the first case, Yoon et al's work~\cite{yoon2018design} does classification of the architectural style of a Minecraft settlement from human-created extracted from the Minecraft Schematics database\footnote{https://www.minecraft-schematics.com}, where humans can submit their creations in Minecraft and tag them with themes such as Medieval, Futurist, Asian, etc. While their approach focuses on classification rather than generation, it is possible that generative approaches using a similar dataset could yield good results.

 Real-world data has also been previously used for procedural map generation, such as in~\cite{barros2015balanced} where geographic data is used to build maps for a turn-based strategy game. Barros et al~\cite{barros2018data} provides a good overview of other real-world data-based pcg methods. A program might take real world data into account when selecting the location of bridges, choosing what type of settlement best fit a certain geography and determining the relative distances between distinct districts or buildings. Metrics derived from real-world data, such as population density and income distribution could also play a role in the creation of a settlement.
 
 While these data-driven approaches come with their own set of challenges, they could conceivably achieve a higher level of adaptability than hand-written rules, and also contribute towards a more realistic feel for the generated settlements.

\subsection{Simulationist Approaches}

Real-life settlements are product of a long and complex development process full of interaction between the settlement's inhabitants, natural and built environment, and other agents. The entries in this competition tried to directly generate the end results of this process instead of simulating the self-organizing process that created the settlements. But there are examples of games, such as Dwarf Fortress \cite{dwarffortress}, that simulate extensive histories to generate their game world. There are also agent-based approaches to environment generation, which to various degrees replicate natural development~\cite{doran2010controlled}. Going further, one could even approach this with an embodied, multi-agent simulation, where simulated agents gather resources, place buildings, travel through and improve the world. This would require a lot more investment in time and effort up front, but could potentially pay off in the long run, because the resulting artifact would not just \textit{look} like but \textit{be} the result of an adaptive process over time.  

\section{New Bonus Challenge: GDMC Chronicle}
Both our evaluation results and the comments made by the judges indicated that the competitors did little to address the challenge of creating an evocative narrative. To counter this we introduced an optional bonus challenge for the second iteration of the GDMC challenge \cite{Salge2019}. That means participants can compete as normal, but those that indicate their participation in the bonus challenge will, in addition, also be able to compete in the Chronicle Challenge. 

The goal of the chronicle challenge is to generate a chronicle, which is a piece of text (written in English) that explicitly tells the story or history of the generated settlement. There is no restriction on the format of the chronicle, but it should be written in a Minecraft book, and placed inside the settlement – ideally in a place where it is easily found. The entries will be evaluated by the same judges evaluating the settlements, and their evaluation is based on two criteria. The overall quality of the chronicle, and how well it fits or relates to the settlement. A good way to illustrate the idea of ``fit'' is to imagine that we will apply the submitted settlement generation algorithms to several maps - and in each case a chronicle should be generated. It should be obvious from each chronicle which of the settlements it belongs to, so if we switched out the chronicles, they would not fit with the other settlements.

The bonus challenge requires textual narrative generation for which there are many existing approaches \cite{gervas2017deconstructing,kybartas2017survey}. But similar to the base challenge, here we are looking for adaptive PCG - where the generated textual content has to fit together with existing content rather than clean slate generation. 

\section{Conclusion}

Looking back at the first year of the GDMC competition we observed different approaches that combined existing methods in novel ways. The current approaches show promise, but there is plenty of room for improvement. In part, this is about catching up or realizing existing, state of the art approaches to city generation that are present in scientific literature \cite{muller2006procedural,parish2001procedural,kelly2007citygen,egs.20091045}. But there are also open challenges, in large part due to the adaptive aspect of this challenge, that have not been addressed anywhere. We are planning to run further iterations of this challenge on a regular basis, at international conferences on Games, AI and/or computational creativity. We have also recently extended our submission options and are now allowing the submission of Java-based settlement generators. We hope that this will make it easier and more attractive for the large and active community of Minecraft modders to participate in the challenge. The competition is explicitly designed to encourage ``citizen science'' and attract participants from outside of academia. We are looking forward to future rounds of the GDMC competition, as we are interested in seeing novel and creative ideas to address the great range of challenges this competition poses.

\begin{acknowledgements}

\todo[inline]{All: Add your funding if required, or other acknowledgements}
CS is funded by the EU Horizon 2020 programme / Marie Sklodowska- Curie grant 705643. RC gratefully acknowledges the financial support from Honda Research Institute Europe (HRI-EU). Michael Cerny Green acknowledges the financial support of the GAANN program. Many thanks to our panel of expert judges who donated their time to evaluate the different entries. Also thanks to Squirrel Eiserloh for providing us with screenshots used in this paper. 
\end{acknowledgements}

\bibliographystyle{spmpsci}      
\bibliography{GDMCbib.bib}  

\newpage
\section{Appendix}

\subsection{Evaluation Criteria}

The following question were given to the judges to illustrate the four evaluation criteria. We explicitly pointed out that these are just example questions, they do not fully define the criteria, and are deliberately kept brief to allow our judges and participants to further interpret the criteria as they see fit:

\subsubsection{Adaptability}
\begin{itemize}
\item Do the structures in the settlement adapt to the terrain?
\item Do the structures in the environment reflect the environment, i.e. usage of available material, adaptation to the biome?
\item Does the settlement take advantage of terrain features or compensate for problems with the terrain?
\item Are the settlements different in reaction to the different initial maps?
\item Are there any other ways in which the settlement adapts to the given maps?
\end{itemize}

\subsubsection{Functionality} 
\begin{itemize}
\item Does the settlement provide protection from danger?
	\begin{itemize}
	\item Does it keep mobs from spawning?
	\item Does it keep mobs out?
	\item Protection from other environmental dangers?
	\end{itemize}
\item Is the settlement accessible to a player avatar in survival mode?
	\begin{itemize}
	\item Can you walk to everywhere?
	\item Does the settlement provide faster modes of transport?
	\item How easy is it to find your way around?
	\end{itemize}
\item Does the settlement provide the player with additional affordances?
\item Does the settlement make resources easy to obtain?
\item Is there an easy way to get food?
\item Does the settlement provide functionality to the villagers?
\item Does the settlement reflect the embodiment of the player avatar?
\item Is it appropriately scaled?
\end{itemize}

\subsubsection{Believable and Evocative Narrative}
\begin{itemize}
\item Is the settlement evoking an interesting story?
\item After looking at the settlement, could you give a short description of what this settlement is about that sets it apart from other settlements?
\item Is it clear what the function of the settlement is?
\item Does this function make sense in regards to the terrain and environment it is in? I.e. is the logging camp in a forest, the harbour town at the sea, … ?
\item Is the functionality of the settlement supporting this narrative function? I.e. does the fortified frontier settlement have functioning walls, is the farming village equipped with functioning fields?
\item Does the final settlement give any indication of how the settlement developed? 
\item Is is possible to look at the settlement and imagine in what order things were built, or what stages the development of the settlement took? 
\item Is there an indication of the history of the settlement evident in the structure?
\item Are there any convincing and consistent allusions to human cultures or specific points in history that the settlement is modeled after
\begin{itemize}
\item Does the settlement have a culture - either fictional or historical - that is evident from the settlement?
\item Do you know things about this culture just by looking at the settlement?
\end{itemize}
\end{itemize}

\subsubsection{Visual Aesthetics}
\begin{itemize}
\item Does the settlement look good?
\item Is there a consistent look to the settlement? Does it appear that all structures belong to the same settlement?
\item Is there an appropriate level of variation in the existing structures?
\item Are there any jarring features that make the settlement look unbelievable?
\end{itemize}

\subsection{Advisory Board}
\label{AB}

At the conception of this competition the organization committee assembled a group of 12 experts to form an advisory board. The advisory board has three different functions. First, at several stages the advisory board had the opportunity to comment on the plans for the competition and give feedback. They gave feedback on the general set up for the competition, the questions used for the evaluation, the overall road map and scientific relevance of the challenge. They were also polled after the evaluation of the first year was finished for a post mortem of the evaluation process. 

Second, they would form the initial pool of judges. That means, in the first year they all received the material to judge the GDMC entries. Third, they were allowed to nominate guest judges who would then serve as additional, equal judges for the specific year. In the first year five guest judges were nominated by members of the AB. So the judgement material was send out to 17 people, of which 8 returned filled in score sheets within the time frame allotted for judging. The following is a list of all advisory board members and all guest judges. We also briefly list the area of expertise that lead us to recruit the specific experts. We as the organizers were grateful for their support and their time. 

\subsubsection{Advisory Board Members}
\begin{itemize}
    \item Cedric de Jacquelot - Minecraft Modder, Millenaire Mod
    \item Mike Cook - Academic, Automatic Game Design, Procedural Content Generation
    \item Andy Nealen - Academic, Game Designer, Computer Graphics, Architect 
    \item Diegeo Perez Liebana - Academic, Game AI, Competition Organizer
    \item Mark R. Johnson - Academic, Game Designer, Procedural Content Generation
    \item Tom Froese - Academic, Artificial Life, Digital Archaeology
    \item Rafael Bidarra - Academic, Computer Graphics, Procedural Content Generation
    \item Jialin Liu - Academic, Game AI, Competition Organizer
    \item Mitu Khandaker - Academic, Game Designer
    \item Jonas Buechel - Urbanist, Participatory City Planning
    \item Richard Bartle - Academic, Game Designer, Design of Virtual Worlds
    \item Gillian Smith - Academic, Procedural Content Generation
\end{itemize}

\subsubsection{Guest Judges}
\begin{itemize}
    \item Mike Green - Academic, AI in Games, Automatic Tutorial Generation for Games, BA in Classical Civilizations
    \item Raluca Gaina - Academic, General Game AI, Reinforcement Learning to play Minecraft Competition Organizer
    \item Ben Kybartas - Academic, PCG in Games, Narrative Generation in Games, Content Orchestration
    \item Mike Preuss - Academic, AI in Games, Search and Evolution
    \item Squirrel Eiserloh - Commercial Game Designer / Programmer, Lecturer on Game Design
\end{itemize}

\section{Supporting Material}

We can provide the following material for publication with the paper:

\begin{itemize}
    \item List of questions to illustrate the evaluation criteria
    \item 3 competition maps in Minecraft map format
    \item 12 minecraft maps with AI generated settlements in Minecraft map format.
    \item Code for the 4 entries, in Python. 
\end{itemize}

We also host all this material on our webpage at http://gendesignmc.engineering.nyu.edu/.

\end{document}